\documentclass[letterpaper, 10 pt, conference]{ieeeconf}  % Comment this line out if you need a4paper

\IEEEoverridecommandlockouts                              % This command is only needed if 
\usepackage{algorithm}
\usepackage[noend]{algpseudocode}
\usepackage{amssymb}
\usepackage{mathtools}

\newtheorem{theorem}{Theorem}

\usepackage{booktabs}

\usepackage{url}
\newcommand{\para}[1]{\noindent \textit{#1}}

\newcommand{\allDrawbells}[0]{D}
\newcommand{\drawbell}[1]{d_{#1}}
\newcommand{\drawVel}[1]{\lambda_{#1}}
\newcommand{\totalDrawVel}[0]{\Lambda}

\newcommand{\drawRatio}[1]{r_{#1}}
\newcommand{\drawRatioV}[0]{R}

\newcommand{\allCrushers}[0]{C}
\newcommand{\crusher}[1]{c_{#1}}

\newcommand{\constantTimeRoadmap}[0]{G}
\newcommand{\allPoses}[0]{Q}
\newcommand{\pose}[1]{q_{#1}}
\newcommand{\allEdges}[0]{E}
\newcommand{\drawbellsAt}[0]{\bar{D}}
\newcommand{\crushersAt}[0]{\bar{C}}
\newcommand{\timestepDuration}[0]{\tau}

\newcommand{\allAg}[0]{A}
\newcommand{\ag}[1]{a_{#1}}
\newcommand{\pickupTime}[0]{T_{pk}}

\newcommand{\depositTime}[0]{T_{dp}}
\newcommand{\poseConflictFamily}[0]{\mathcal{C}^Q}
\newcommand{\edgeConflictFamily}[0]{\mathcal{C}^E}
\newcommand{\allAgAt}[1]{\boldsymbol{\pi}^{#1}}
\newcommand{\agAt}[1]{\pi_{#1}}
\newcommand{\allAgHas}[1]{\boldsymbol{\sigma}^{#1}}
\newcommand{\agHas}[1]{\sigma_{#1}}

\newcommand{\ct}[0]{T}
\newcommand{\agPerm}[0]{\Omega}

\newcommand{\OSBCMInstance}[1]{\mathcal{I}_{#1}}
\newcommand{\validPlanSet}[0]{\Xi}
\newcommand{\deposits}[0]{\textsc{deposits}}

\newcommand{\allShovelSt}[0]{\Phi}
\newcommand{\shovelSt}[0]{\phi}

\newcommand{\atT}[0]{\boldsymbol{at}}
\newcommand{\at}[0]{at}
\newcommand{\mvT}[0]{\boldsymbol{mv}}
\newcommand{\mv}[0]{mv}
\newcommand{\pkT}[0]{\boldsymbol{pk}}
\newcommand{\pk}[0]{pk}
\newcommand{\dpT}[0]{\boldsymbol{dp}}
\newcommand{\dpF}[0]{dp}

\newcommand{\adj}[0]{\textsc{adj}}
\newcommand{\accessibleFrom}[0]{\bar{Q}}

\newcommand{\convFLIMM}[0]{\textsc{convFLIMM}}
\newcommand{\nextAgState}[0]{\textsc{nxtState}}
\newcommand{\bestAllAgAt}[0]{\boldsymbol{\pi}^*}
\newcommand{\bestAllAgHas}[0]{\boldsymbol{\sigma}^*}
\newcommand{\bestAgPerm}[0]{\Omega^*}
\newcommand{\SAMM}[0]{\textsc{SAMM}}
\newcommand{\FLIMM}[0]{\textsc{FLIMM}}

\newcommand{\bestCt}[0]{T^*}

\newcommand{\exeDrawRate}[0]{z_i(t)}

\newcommand{\SAMMS}[0]{\textsc{SAMMS}}
\newcommand{\partitionDrawbells}[0]{\textsc{partitionDrawbells}}
\newcommand{\allFamilies}[0]{\mathcal{D}}
\newcommand{\family}[1]{D'_{#1}}
\newcommand{\allAgAtF}[1]{\boldsymbol{\pi}^{(#1)}}
\newcommand{\allAgHasF}[1]{\boldsymbol{\sigma}^{(#1)}}
\newcommand{\agPermF}[1]{\Omega^{(#1)}}
\newcommand{\genReferenceSubcycle}[0]{\textsc{genRefSubcycle}}
\newcommand{\agAtSt}[0]{\bar{\boldsymbol{\pi}}}
\newcommand{\agHasSt}[0]{\bar{\boldsymbol{\sigma}}}
\newcommand{\genSubcycle}[0]{\textsc{genSubcycle}}
\newcommand{\composeSubcycles}[0]{\textsc{composeSubcycles}}
\newcommand{\FSSAMM}[0]{\textsc{FS-SAMM}}

\newcommand{\reducedBCMInstance}[1]{\mathcal{I}'_{#1}}
\newcommand{\referenceDrawbell}[1]{\psi_{#1}}
\newcommand{\referenceDrawRatio}[1]{\rho_{#1}}

\newcommand{\reducedDrawRatioV}[1]{R'_{#1}}

\newcommand{\allSubcycles}[0]{\Xi}
\newcommand{\subcycle}[1]{\xi_{#1}}
\newcommand{\frequency}[1]{f_{#1}}
\newcommand{\pickup}[0]{\textsc{pickup}}
\newcommand{\executionRatio}[1]{\eta_{#1}}
\newcommand{\allExecutionRatios}[0]{\boldsymbol{\eta}}
\newcommand{\run}[0]{\textsc{run}}

\newcommand{\reduceBCM}[0]{\textsc{reduceBCM}}

\newcommand{\agAtStF}[1]{\bar{\pi}_{#1}}
\newcommand{\agHasStF}[1]{\bar{\sigma}_{#1}}

\newcommand*\circled[1]{\raisebox{.5pt}{\textcircled{\raisebox{-.9pt} {#1}}}}

\newcommand{\timer}[0]{timer}

\newcommand{\nullVal}[0]{\textsc{null}}

\title{\LARGE Eventually Optimal and Scalable Multi-Agent Planning\\ for Block Cave Mining}

\author{Christopher Leet$^{1}$, Paolo Forte$^{2}$, Uwe Köckemann$^{2}$, Henrik Andreasson$^{2}$, Sven Koenig$^{1,2}$% <-this % stops a space
\thanks{This work was partially supported by the European Union’s Horizon Europe Framework Programme under grant agreement No 101070596 (euROBIN). \emph{(Corresponding author: Christopher Leet)}}
\thanks{© 2026 IEEE. Personal use of this material is permitted. Permission from IEEE must be obtained for all other uses, in any current or future media, including reprinting/republishing this material for advertising or promotional purposes, creating new collective works, for resale or redistribution to servers or lists, or reuse of any copyrighted component of this work in other works.}% <-this % stops a space
\thanks{$^{1}$University of Southern California Los Angeles, California, USA {\tt\small \{cjleet,skoenig\}@usc.edu}}%
\thanks{$^{2}$Centre for Applied Autonomous Sensor Systems, Örebro University, Örebro, Sweden {\tt\small <name>.<surname>@oru.se}}
}

\begin{document}

\maketitle
\thispagestyle{empty}
\pagestyle{empty}

%%%%%%%%%%%%%%%%%%%%%%%%%%%%%%%%%%%%%%%%%%%%%%%%%%%%%%%%%%%%%%%%%%%%%%%%%%%%%%%%
\begin{abstract}

Automation in underground mining has the potential to significantly enhance safety, operational efficiency, and sustainability. However, effectively coordinating fleets of autonomous vehicles in dynamic mine environments introduces substantial challenges in both optimization and motion planning. To address these challenges, we introduce and formalize the \emph{Block Cave Mining (BCM)} problem, which focuses on computing a transport plan that maximizes ore throughput while satisfying draw ratio constraints. To solve this problem, we propose SAMM, an eventually optimal anytime solver that jointly integrates task assignment, scheduling, and path planning via a mixed-integer linear programming formulation. To improve scalability, we also introduce SAMMS, a variant of SAMM that trades optimality guarantees for efficiency by decomposing the problem into shorter planning subcycles. %and CAMM, which leverages a coarse actor-based decomposition to improve scalability. 
Experimental evaluations using realistic industrial mine scenarios demonstrate that SAMMS achieves near-optimal throughput and scales effectively to larger fleets and mine layouts.

\end{abstract}

%%%%%%%%%%%%%%%%%%%%%%%%%%%%%%%%%%%%%%%%%%%%%%%%%%%%%%%%%%%%%%%%%%%%%%%%%%%%%%%%

\section{Introduction}
\begin{figure}[!ht]
    \centering
    \includegraphics[width=0.9\linewidth]{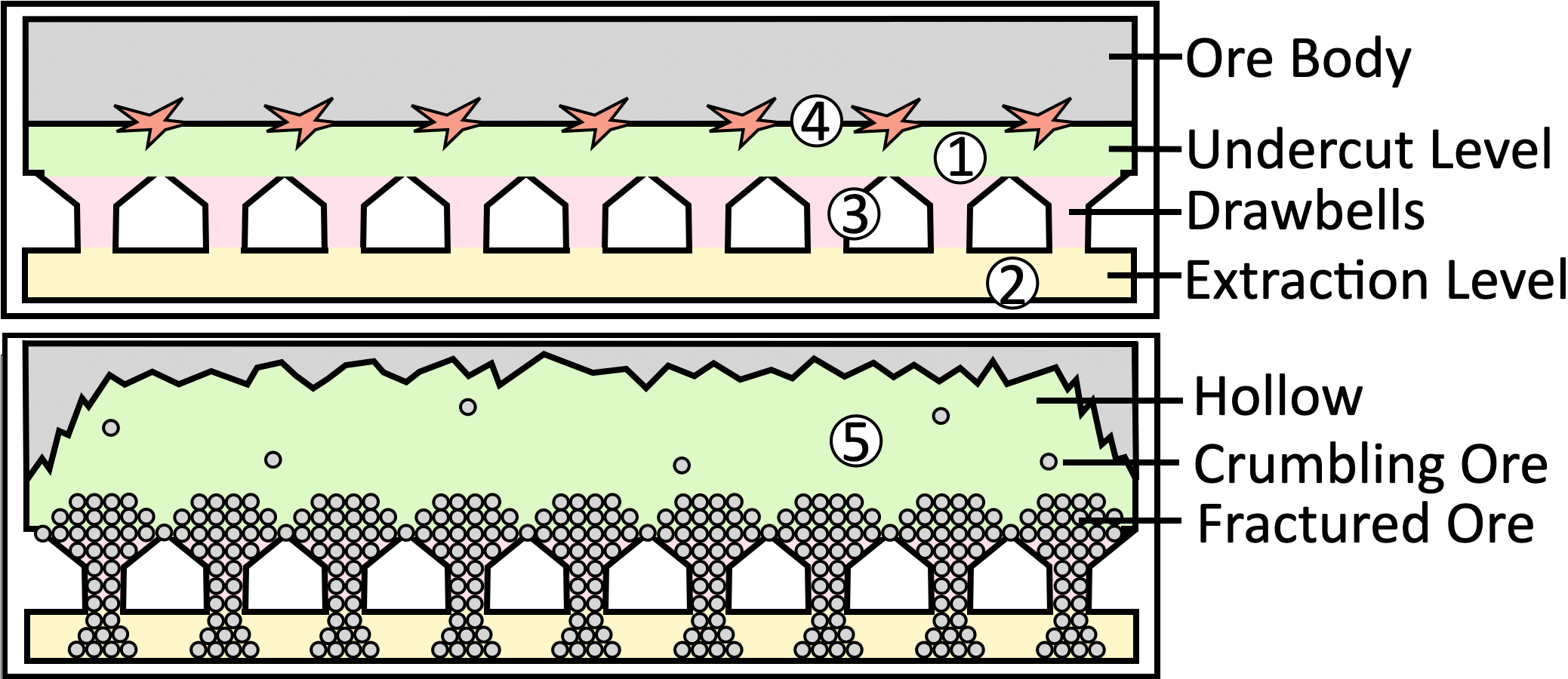}
    \vspace{-3mm}
    \caption{A side view of the block cave mining process. Top: building the infrastructure. Bottom: extracting ore.}
    \label{fig:block_cave_mining}
\end{figure}

Autonomous material transport systems are becoming increasingly important to mining~\cite{automation_in_mining}, as  they can improve efficiency, safety, and reduce costs~\cite{rio_tinto_automation}. Among the mining methods that benefit from automation is block cave mining (see Fig.~\ref{fig:block_cave_mining}), a mining technique used to extract large volumes of ore from underground mines~\cite{block_cave_mining}. It begins with the excavation of an undercut layer $\circled{1}$ and extraction tunnels $\circled{2}$ beneath an ore body. Vertical drawbells $\circled{3}$ connect the undercut layer to the extraction tunnels. The undercut layer is then blasted $\circled{4}$ causing the base of the ore body to fracture and collapse. The ore is transported from the drawbells  to crushers which pulverize it and deposit it into containers for removal. As ore is extracted, a hollow $\circled{5}$ forms at the base of the ore body. This loss of support causes the remainder of the ore body to slowly crumble under its own weight and fall into the drawbells for collection.

To automate the ore collection process, a block cave mine operator must generate a plan that specifies how autonomous agents transport ore to crushers. 
%To keep the debris pile stable, the rate at which ore is extracted from each drawbell must be a fixed percentage of the total extraction rate. 
This production plan is to alleviate stress level build up in the mine as well as fragmentation of the ore, the amount of material drawn from each drawbell is important to keep within bounds and changes slowly over time~\cite{Guest2006AnAO}. In this work we assume the rate at which ore is extracted is a fixed percentage of the total extraction rate (if the rates are updated we treat it as a new problem). 
%Since the pile is usually asymmetric, these percentages vary between drawbells. 
These percentages vary between drawbells, for example, one drawbell might contribute 7\% of the extracted ore while another might only contribute 1\%. We define the task of finding a transport plan which maximizes the overall extraction rate as the  \emph{Block Cave Mining (BCM)} problem. The BCM problem presents three challenges. First, since agents must haul ore for an indefinite amount of time, a solver must construct an arbitrarily long plan. Second, the BCM problem is composed of three, tightly coupled, components:  task assignment (which drawbell to visit), scheduling (when to visit) and path planning (how to get there). Finally, BCM agents have heavy restrictions on their movement. Most prior work on \emph{Multi-Agent Path Finding (MAPF)} models the environment as a 2D grid graph with holes and assumes that agents can move in any cardinal direction in constant time. BCM agents, however, are bulky, center articulated vehicles only slightly narrower than the mine's tunnels. Even basic actions such as turning around require a complex series of maneuvers at a junction, complicating path planning.

To the best of our knowledge, no optimal solution to the BCM problem exists. Current solutions use heuristics such as statically assigning agents to fixed subsets of drawbells, which limits optimality. To bridge this gap, we introduce SAMM (Simple Anytime Mine MAPF), an eventually optimal method designed to meet the challenges above. SAMM achieves continuous operation by producing cyclic plans, plans that can be looped because they start and end with the team of agents in the same configuration. It avoids the pitfalls of decomposition by casting task assignment, scheduling and path planning into a single \emph{Mixed Integer Linear Program (MILP)} which captures their inter-dependencies and decides the optimal number of agents for a given cycle length. Finally, to model maneuvering in confined tunnels, SAMM constructs a state graph over discrete poses (position–orientation pairs), where transitions encode feasible motion primitives executable in constant time. 
The industrial applicability of  MILP approaches, such as SAMM, to MAPF is limited since they typically do not scale well. 

Therefore, we propose an additional, scalable solver, SAMMS (SAMM with Subcycles). The difficulty of generating a cyclic plan using a ILP grows exponentially with the plan’s length. Instead of generating one long cycle which services every drawbell, SAMMS generates a sequence of subcycles, each of which services a subset of drawbells.  Each subcycle starts and ends in the same state, allowing them to be seamlessly concatenated. SAMMS constructs a transport plan online by concatenating these subcycles. Whenever a subcycle completes, SAMMS selects a new subcycle to execute. We prove that SAMMS is always asymptotically feasible and converges rapidly. The longer it runs, the closer it approximates each drawbell's targeted draw rate.

We evaluate SAMM and SAMMS on benchmark scenarios derived from real-world mine layouts and demonstrate their ability to solve industrial scale BCM instances.

\section{Related Work}
\emph{Multi-Agent Path Finding (MAPF)}~\cite{Stern2019MultiAgentPD} is the problem of computing collision-free paths for multiple agents in a shared environment. %The MAPF problem has been solved via prioritized planning~\cite{PP}, search~\cite{CBS}, and answer set programming~\cite{ASP_LMAPF_offline} among other methods. 
MAPF has been solved in different industrial domains, including warehouse logistics~\cite{Agaskar2025DeepFleetMF}, transportation systems~\cite{Li2023IntersectionCW}, and construction~\cite{rl_CCP}. 
Classical MAPF solutions mainly rely on two main paradigms: search-based and compilation-based approaches. Search-based methods emphasize solution optimality by employing graph search, heuristics, and specialized techniques to minimize collisions and search overhead. Examples include prioritized planning~\cite{PP}, conflict-based search~\cite{CBS}, and priority-based search~\cite{Chan2023GreedyPS}. Compilation-based approaches reformulate MAPF as an optimization problem, such as Boolean satisfiability~\cite{MAPF_SAT}, mixed-integer linear programming~\cite{Lam2019BranchandCutandPriceFM}, or answer set programming~\cite{ASP_LMAPF_offline}. While these methods can yield high-quality or even provably optimal solutions, they often face scalability issues on larger instances. More recently, the field has witnessed an increasing interest in learning-based methods, including reinforcement learning~\cite{rl_CCP} and foundation model driven approaches~\cite{Agaskar2025DeepFleetMF}. However, most existing MAPF algorithms make simplifying assumptions that do not hold in block cave mining. They generally assume grid-based navigation, a fixed number of agents, and that task assignment and scheduling are precomputed, focusing solely on path planning. Our approach jointly optimizes the number of agents, task assignment, scheduling, and path planning, and produces plans that can be executed continuously.

Various approaches have been explored for underground mine planning and scheduling~\cite{Khodayari2015MathematicalPA}. Production scheduling methods can be broadly categorized into heuristic methods and exact optimization techniques.
Heuristic approaches aim to generate near-optimal solutions within a reasonable time, particularly when identifying the optimal solution under complex constraints is computationally infeasible. Examples include block aggregation heuristics~\cite{Tabesh2011TwostageCA} and optimization-based decomposition heuristics~\cite{OSullivan2015OptimizationbasedHF}. While computationally efficient, these methods cannot guarantee globally optimal schedules.
Conversely, exact mathematical programming methods have been widely applied to block-caving production scheduling, as they allow systematic exploration of feasible solutions while accounting for operational constraints and providing bounds on optimality. Techniques such as linear programming~\cite{Guest2006AnAO},  and quadratic programming~\cite{Donaldson2020OperationalAR} have been proposed to optimize production plans. However, these approaches neither generate optimal transport plans nor scale.

%Beyond the algorithmic core, real-world manufacturing systems pose integration challenges. The warehouse servicing problem considers the coordinated transport of goods by mobile agents within strict time constraints, involving joint task assignment, scheduling, and path planning. Similarly, the smart factory embedding problem addresses the joint optimization of production processes and transport logistics. Recent solvers, such as  the anytime cyclic embedding solver~\cite{ACES}, are designed to meet these demands at industrial scale, enabling continuous, adaptive decision-making in smart factories \cite{CCPP}. Algorithmic solutions for Load-Haul-Dump (LHD) fleet management often focus on the integration problem of scheduling and path planning. The problem is not just determining where an agent should go next (dispatching), but also which path it should take and when it should travel, as paths may be temporarily blocked by other agents moving in the opposite direction. These problems are often solved using shortest-path algorithms on a graph-based representation of the mine's haulage network. However, unlike a simple road map, the edge costs in this graph are dynamic, changing in real-time to reflect traffic congestion, vehicle orientation (LHDs can move forward and backward), and other operational constraints \cite{Gamache2002FleetMS}.

\section{Problem Formulation}
We now formally present the block cave mining problem.
% In this section, we first introduce key concepts foundational to the remainder of the paper, some illustrated in Fig.~\ref{fig:small_mine}, and then formally define the One-Shot Block Cave Mining (OS-BCM) problem.

\subsection{Mine}
A block cave mine is a collection of drawbells and ore crushers connected by a network of narrow tunnels.

\para{Drawbells}. A drawbell is a shaft filled with crushed ore. Let $\allDrawbells{} := \{\drawbell{1}, \drawbell{2}, \ldots\}$ be the set of drawbells in a mine. Let draw rate $\drawVel{i}$ be the average rate at which ore is extracted from drawbell $\drawbell{i}$. Each drawbell's draw rate is coupled to the draw rates of the other drawbells. Let $\totalDrawVel{} := \sum_{\drawbell{i} \in \allDrawbells{}} \drawVel{i}$ be the total ore extraction rate. Each drawbell $\drawbell{i}$’s extraction rate must be a fixed fraction $\drawRatio{i} := \drawVel{i} / \totalDrawVel{}$ of the total. Let $\drawRatio{i}$ be the drawbell $\drawbell{i}$'s draw ratio, and the vector of draw ratios $\drawRatioV{} := \langle \drawRatio{1}, \drawRatio{2}, \ldots \rangle$ be
the mine’s draw ratio configuration.

\para{Crushers.} A crusher grinds fractured ore and then deposits it in a container for extraction. Let $\allCrushers{} := \{\crusher{1}, \crusher{2}, \ldots\}$ be the set of crushers in the mine. Ore can be transported from any drawbell to any crusher.

\para{Constant Time Roadmap.} We model the routes that agents in the mine can follow using a Constant Time Roadmap (CTR), i.e., a directed graph $\constantTimeRoadmap{} := (\allPoses{}, \allEdges{})$ where each vertex $\pose{i} \in \allPoses{}$ represents a pose (a position and orientation pair). There is an edge $(\pose{i}, \pose{j}) \in \allEdges{}$ iff an agent can move from pose $\pose{i}$ to pose $\pose{j}$ within $\timestepDuration{}$ seconds and without passing through any other pose. Time in a block cave mine is discretized. One timestep lasts $\timestepDuration{}$. Thus an agent can always move from any pose to any adjacent pose in exactly 1 timestep.

A drawbell and a crusher is accessible from a pose $\pose{i}$ if an agent at $\pose{i}$ can load ore into it and unload ore from it. Let $\drawbellsAt(\pose{i}) \subseteq \allDrawbells{}$ and $\crushersAt(\pose{i}) \subseteq \allCrushers{}$ denote the drawbells and crushers serviceable from $\pose{i}$. A valid CTR has the following properties: (1) The roadmap is strongly connected. If the roadmap is disconnected, it may be impossible to carry ore from some drawbell to a crusher; and (2) each drawbell/crusher must be accessible from at least one pose, that is: $\allCrushers{} = \bigcup_{\pose{i} \in \allPoses} \crushersAt(\pose{i})$ and $ \allDrawbells{} = \bigcup_{\pose{i} \in \allPoses} \drawbellsAt(\pose{i})$.
%
%\begin{enumerate}
    %\item The roadmap is strongly connected. If the roadmap is disconnected, it may be impossible to carry ore from some drawbell to a crusher.
    %\item  Each drawbell/crusher must be accessible from at least one pose: $\allCrushers{} = \bigcup_{\pose{i} \in \allPoses} \crushersAt(\pose{i}) \wedge \allDrawbells{} = \bigcup_{\pose{i} \in \allPoses} \drawbellsAt(\pose{i})$.
%\end{enumerate}

%\para{Property 1)} The CTR is required to be strongly connected.

%\para{Property 2)} A drawbell or crusher is accessible from a pose $\pose{i}$ if an agent at $\pose{i}$ can pick up ore from or deposit ore into it. Each drawbell and crusher must be accessible from at least one pose $\pose{i} \in \allPoses{}$. Let $\drawbellsAt{}(\pose{i}) \subseteq \allDrawbells{}$ and $\crushersAt{}(\pose{i}) \subseteq \allCrushers{}$ be the sets of drawbells and crushers that can be serviced from pose $\pose{i}$. 

% This requirement can be formalized as:

% \begin{equation*}
% \bigcup_{\pose{i} \in \allPoses{}} \drawbellsAt{}(\pose{i}) = \allDrawbells{} \; \mbox{and} \; \bigcup_{\pose{i} \in \allPoses{}} \crushersAt{}(\pose{i}) = \allCrushers{}.
% \end{equation*}

%\para{Property 3)} An agent must be able to move between any pair of adjacent poses $(\pose{i}, \pose{j}) \in \allEdges{}$ during a fixed time interval $\timestepDuration{}$. We discretize time so that $\timestepDuration{} = 1$.

\subsection{Agents} 
A team of agents $\allAg{} := \{\ag{1}, \ag{2}, \ldots\}$ transports ore between drawbells and crushers. Each agent is an autonomous vehicle. An agent can perform four types of actions:

\para{Wait.} The agent idles in its current pose for one timestep.

\para{Move.} The agent moves to an adjacent pose in the roadmap.

\para{Pick Up}. The agent picks up ore from a drawbell $\drawbell{j} \in \drawbellsAt{}(\pose{i})$ accessible from its current pose $\pose{i}$.

\para{Deposit.} The agent deposits ore into a crusher $\crusher{j} \in \crushersAt{}(\pose{i})$ accessible from its current pose $\pose{i}$.

Each agent is equipped with a shovel that has a capacity of exactly one ore unit. Shovels cannot be partially filled, and each pickup or deposit action transfers precisely one unit between a drawbell and the agent, or the agent and a crusher. Picking up ore from drawbell $\drawbell{j}$ takes $\pickupTime{}$ timesteps, and depositing it into crusher $\crusher{j}$ takes $\depositTime{}$ timesteps. If an agent at pose $\pose{i}$ begins a pickup or deposit action, it occupies pose $\pose{i}$ until that action is completed. The state of an agent $\ag{i}$ at timestep $t$ is  a pair $(\agAt{it}, \agHas{it})$ where $\agAt{it} \in \allPoses{}$ denotes the agent's pose and $\agHas{it}$ denotes the state of its shovel (0 for empty, and 1 for full).

\subsection{Collision}
Two poses $\pose{i}, \pose{j} \in \allPoses{}$ conflict iff two agents cannot occupy them simultaneously without colliding. We represent pose conflicts with a pose conflict family $\poseConflictFamily{} \subseteq 2^\allPoses{}$. Each element of $\poseConflictFamily{}$ is a pose conflict set $S$, a set of pairwise conflicting poses. Any set of pairwise conflicting poses $S' \subseteq \allPoses{}$ is a subset of some pose conflict set $S \in \poseConflictFamily{}$. A pose conflicts with itself. If a pose $\pose{i}$ does not conflict with any other pose, the singleton set $\{\pose{i}\}$ is contained in $\poseConflictFamily{}$ to represent $\pose{i}$'s self-conflict. Consequently, every pose appears in a set in $\poseConflictFamily{}$ at least once.

Similarly, two edges $(\pose{h}, \pose{i}), (\pose{j}, \pose{k}) \in \allEdges{}$ conflict iff two agents cannot traverse them simultaneously without colliding. We represent edge conflicts with an edge conflict family $\edgeConflictFamily{} \subseteq 2^\allEdges{}$. Each element of $\edgeConflictFamily{}$ is an edge conflict set of pairwise conflicting edges. Any set $S$ of pairwise conflicting edges in $\allEdges{}$ is a subset of some edge conflict set in $\edgeConflictFamily{}$. To prevent collisions, no two agents may occupy poses in the same pose conflict set in $\poseConflictFamily{}$ at the same timestep or move along edges in the same edge conflict set in $\edgeConflictFamily{}$ from the same timestep to the next one. 

\subsection{Transport Plan}
 
A transport plan $(\allAgAt{}, \allAgHas{})$ describes how agents move ore from drawbells to crushers. It defines the state $(\agAt{it}, \agHas{it})$ of each agent $\ag{i} \in \allAg{}$ at every timestep $t$. As ore must be transported continuously, we generate a cyclic transport plan that can be repeated indefinitely.
A cyclic transport plan is associated with a cycle time $\ct{} \in \mathbb{N}$ and an agent permutation $\agPerm{} : \allAg{} \rightarrow \allAg{}$. The transport plan runs from $t = 0$ to $t = \ct{}$. The state of each agent $\ag{i} \in \allAg{}$ at $t = 0$ must be the same as the state of agent $\agPerm{}(\ag{i})$ at $t = \ct{}$, that is,
\begin{equation*}
    \forall\ \ag{i} \in \allAg{}, (\agAt{i0}, \agHas{i0}) = (\agAt{jT}, \agHas{jT})\ \text{for}\ a_j = \Omega(a_i). 
\end{equation*}
A cyclic transport plan is repeated indefinitely by executing it from $t = 0$ to $t = \ct{}$, relabeling its agents according to $\agPerm{}$, and then repeating the process. A cyclic transport plan is valid iff: (1) it can be realized by assigning each agent a sequence of wait, move, pickup and deposit actions, and (2) no two agents occupy conflicting poses or move along conflicting edges at any time $t$.

\subsection{The Block Cave Mining (BCM) Problem}

A BCM problem can be formalized as a 5-tuple $\OSBCMInstance{} := (\allPoses{}, \allEdges{}, \allDrawbells{}, \allCrushers{}, \drawRatioV{})$, where $(\allPoses{}, \allEdges{})$ is a constant time roadmap, $\allDrawbells{}$ and $\allCrushers{}$ are its drawbells and crushers, and $\drawRatioV{}$ is a draw ratio configuration. The throughput of a cyclic transport plan $(\allAgAt{}, \allAgHas{})$ with cycle length $\ct{}$ is the average units of ore deposited per timestep. If $\deposits{}(\allAgAt{}, \allAgHas{})$ is the number of deposit actions in the plan $(\allAgAt{}, \allAgHas{})$, the plan's throughput is $\deposits{}(\allAgAt{}, \allAgHas{})/\ct{}$. In the Block Cave Mining problem, we are given an BCM instance $\OSBCMInstance{}$ and asked to find a valid cyclic transport plan that maximizes throughput. Let $\validPlanSet{}(\OSBCMInstance{}, \ct{})$ be the set of valid transport plans for instance $\OSBCMInstance{}$ with cycle length $\ct{}$. The BCM is formulated as:
\begin{equation*}
   \max_{\ct{} \in \mathbb{N}, (\allAgAt{}, \allAgHas{}) \in \validPlanSet{}(\OSBCMInstance{}, \ct{})} \frac{\deposits{}(\allAgAt{}, \allAgHas{})}{\ct{}}.
\end{equation*}

\section{Block Cave Mine Benchmarks}

In this section, we describe a family of benchmark block cave mine layouts~\cite{Gomez2020LayoutsBlockCaving}. This family illustrates the layout of real-world block cave mines, composed of three components: drawbell tunnels, crusher tunnels, and a main tunnel.

\para{Drawbell Tunnels.} Drawbell tunnels are linear, dead-end tunnels with drawbells positioned at regular intervals along both sides. These tunnels are narrow, preventing agents from turning around and restricting the drawbells that they can access. In our mine layout, the drawbells on one side of a tunnel can only be accessed by agents facing its entrance, while the drawbells on the opposite side can only be accessed by agents facing its dead end. Consequently, an agent must reverse into a tunnel to access half of its drawbells.

\begin{figure}[t]
    \centering
    \includegraphics[width=0.85\linewidth]{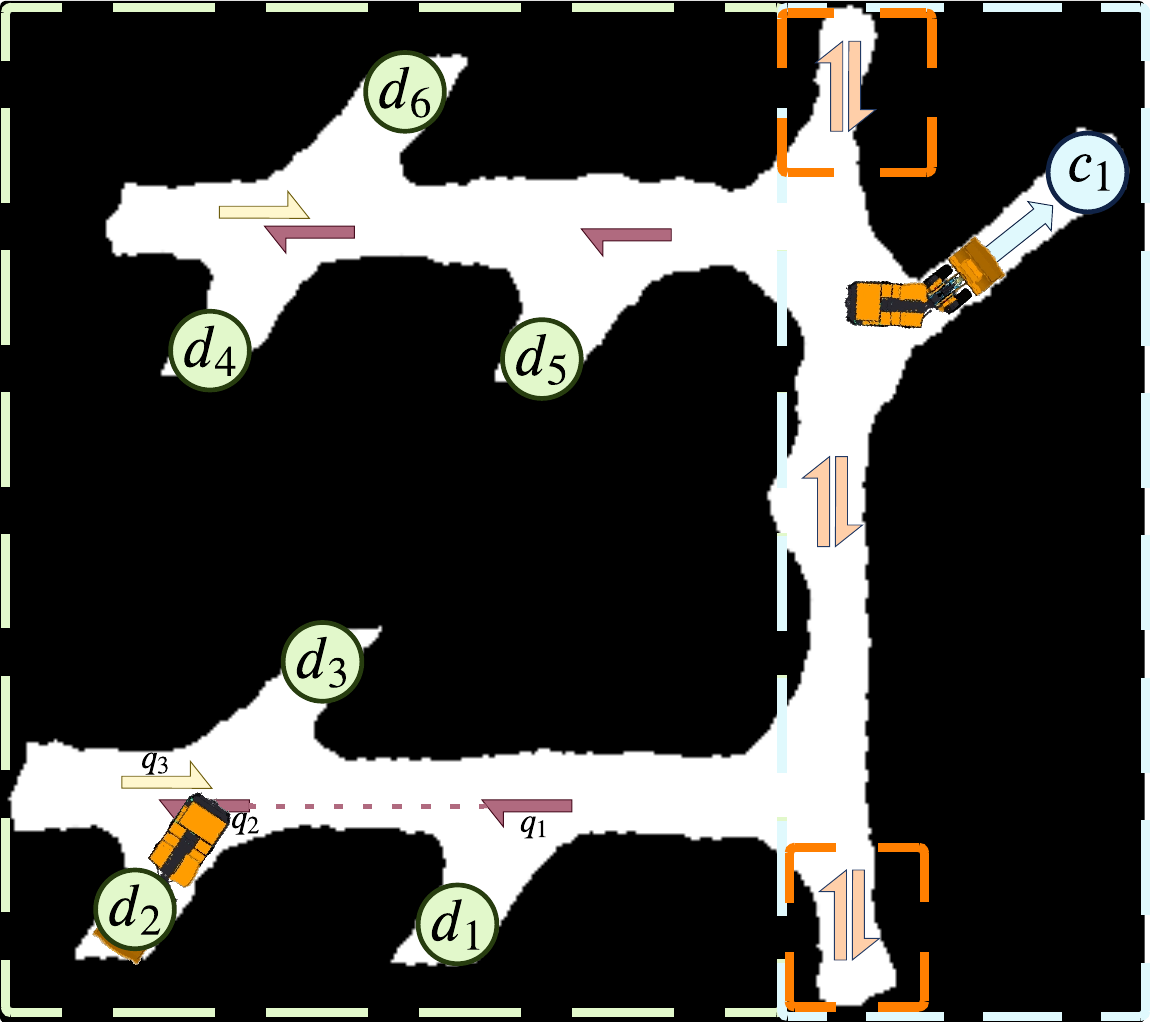}
    \vspace{-3mm}
    \caption{An example mine. Drawbell tunnels are circled in green; the main tunnel and crusher tunnel are circled in blue. Vehicles cannot turn around without complicated maneuvers, facilitated by the poses circled in orange.}
    \label{fig:small_mine}
\end{figure}

There is a bijection between the poses and the drawbells in a drawbell tunnel. Each pose lies directly outside its associated drawbell. Each drawbell is only accessible  from its associated pose. The transitions between these poses form two simple paths: one for poses oriented toward the entrance and the other for poses oriented toward the dead end. Because drawbell tunnels are narrow, an agent following one path cannot pass an agent following the other. Similarly, an agent picking up ore from a drawbell blocks traffic from moving past it.
Fig.~\ref{fig:small_mine} depicts a small mine which has two drawbell tunnels and eight drawbells. The paths of poses oriented towards the entrance and dead end of each drawbell tunnel are shown in red and yellow.

\para{Main Tunnel.} In a block cave mine, the drawbell tunnels run parallel to one another. Each drawbell tunnel connects to a main tunnel which runs perpendicular to them. Two paths of poses oriented in opposite directions run the length of the main tunnel. An agent on one of these paths cannot pass an agent on the other. Each path of poses in a drawbell tunnel connects to both paths of poses in the main tunnel. Fig.~\ref{fig:small_mine} depicts the poses in the mine's main tunnel in orange.

\para{Crusher Tunnels.} Crusher tunnels are short passages that terminate at a crusher. Each crusher tunnel contains a single pose that provides access to its crusher. The three-way junction at the mouth of each drawbell and crusher tunnel allows agents to reorient. The mine in Fig.~\ref{fig:small_mine} has a single crusher tunnel, whose pose is shown in blue.

 We present a benchmark generatorfor block cave mines based on the layout described above. It allows for systematic variation of the number and density of drawbell tunnels, drawbells per tunnel, and crusher tunnels.

\section{The Simple Anytime Mine MAPF Solver}

In this section, we introduce SAMM (the Simple Anytime Mine MAPF), an anytime BCM solver, and prove that it is eventually optimal. We propose a way to accelerate SAMM, and prove that it remains eventually optimal.

We construct SAMM as follows. Let the fixed cycle length (F-BCM) BCM problem  be the problem of finding the best solution to a BCM instance $\OSBCMInstance{}$ which has cycle length $\ct{}$. We introduce FLIMM (Fixed Length ILP-based Mine MAPF),a solver designed to find the optimal solution to the F-BCM. SAMM solves the BCM by running FLIMM with progressively higher values of $\ct{}$ until time runs out.

\subsection{The Fixed Length ILP-based Mine MAPF Solver}

FLIMM solves the F-BCM problem by formulating it as an Integer Linear Program (ILP) as follows.

\para{Variables.} FLIMM's ILP introduces the following variables. Let $\allShovelSt{} \in \{0, 1\}$ be set of possible shovel states and $\shovelSt{} \in \allShovelSt{}$ be an arbitrary shovel state.

\para{Occupancy Tensor.} The tensor $\atT{}$ is a $|\allPoses{}| \times |\allShovelSt{}| \times \ct{}$ tensor. The field $\at{}(\pose{i}, \shovelSt{}, t) \in \{0, 1\}$ indicates if an agent with shovel state $\shovelSt{}$ is at pose $\pose{i}$ at timestep $t$.

\para{Movement Tensor.} The tensor $\mvT{}$ is a $|\allEdges{}| \times |\allShovelSt{}| \times \ct{}$ tensor. The field $\mv{}(
\pose{i}, \pose{j}, \shovelSt{}, t) \in \{0,1\}$ indicates if an agent with shovel state $\shovelSt{}$ moves from pose $\pose{i}$ to pose $\pose{j}$ on timestep $t$. 

\para{Pickup and Deposit Tensors.} The tensors $\pkT{}$ and $\dpT{}$ are $|\allPoses{}| \times |\allDrawbells{}| \times \ct{}$ and $|\allPoses{}| \times |C| \times \ct{}$ tensors. The fields $\pk{}(\pose{i}, \drawbell{j}, t)$ and $\dpF{}(\pose{i}, \crusher{j}, t)$ indicate if an agent at pose $\pose{i}$ picks up ore from drawbell $\drawbell{j}$ and deposits it in crusher $\crusher{j}$ on timestep $t$.

\para{Objective Function.} The objective is to maximize the rate at which ore is removed from the mine.  Since the ratio between each draw rate $\drawVel{i} \in \{\drawbell{i}\}_{i=2}^{|\allDrawbells{}|}$ and $\drawVel{1}$ is fixed, this is equivalent to maximizing the rate at which ore is picked up from drawbell $\drawbell{1}$. Let  $\accessibleFrom{}_i$ be the set of poses from which a drawbell $\drawbell{i} \in \allDrawbells{}$ can be accessed.
\begin{equation*}
\max\sum_{\mathclap{\pose{j} \in \accessibleFrom{}_1}} \ \sum_{t=0}^{\ct{}-1} \pk{}(\pose{j}, \drawbell{1}, t)
\end{equation*}

% \para{Constraints.} There are four categories of constraints: action selection, next state, collision avoidance, and draw ratio constraints. 

\para{Action Selection Constraints.} These constraints limit the actions that an agent may take in a given state. Let $\adj{}(\pose{i})$ denote the set of poses adjacent to pose $\pose{i}$ in the CTR.

\para{Constraint 1.} An agent at pose $\pose{i}$ on timestep $t$ with an empty shovel must move to an adjacent pose $\pose{j} \in \adj{}(\pose{i})$, wait at $\pose{i}$, or pick up ore from a drawbell $\drawbell{j} \in \drawbellsAt{}(\pose{i})$ accessible from pose $\pose{i}$. Every pose in the CTR has a self-loop edge since an agent at pose $\pose{i}$ can always reach $\pose{i}$ in $\timestepDuration{}$ time. The set of variables $\{\mv{}(\pose{i}, \pose{j}, 0, t)\}_{\pose{j} \in \adj{}(\pose{i})} \}$ thus represents both move actions and wait actions. 
\begin{align*}
\forall\ \pose{i} \in &\ \allPoses{}, \forall\ t \in [0..\ct{}-1],\\%
&\at{}(\pose{i}, 0, t) = \sum_{\mathclap{\pose{j} \in \adj{}(\pose{i})}} \mv{}(\pose{i}, \pose{j}, 0, t)  + \sum_{\mathclap{\drawbell{j} \in \drawbellsAt(\pose{i})}} \pk{}(\pose{i}, \drawbell{j}, t). 
\end{align*}

\para{Constraint 2.} An agent at pose $\pose{i}$ on timestep $t$ with a full shovel  must move to an adjacent pose, wait at pose $\pose{i}$, or deposit ore into a crusher accessible from $\pose{i}$. 
\begin{align*}
\forall\ \pose{i} \in & \ \allPoses{}, \forall\ t \in [0..\ct{}-1],\\%
&\at{}(\pose{i}, 1, t) = \sum_{\mathclap{\pose{j} \in \adj{}(\pose{i})}} \mv{}(\pose{i}, \pose{j}, 1, t) + \sum_{\mathclap{\crusher{j} \in \crushersAt{}(\pose{i})}} \dpF{}(\pose{i}, \crusher{j}, t).
\end{align*}

\para{Constraint 3.} FLIMM models a $\pickupTime{}$ timestep pickup action as a one timestep pickup action followed by $\pickupTime{} - 1$ one timestep wait actions. An agent picking up ore from a drawbell $\drawbell{j} \in \drawbellsAt{}(\pose{i})$ at pose $\pose{i}$  on timestep $t$ must therefore wait at $\pose{i}$ for the next $\pickupTime{} - 1$ timesteps.
\begin{align*}
&\forall\ \pose{i} \in \allPoses{},\ \forall\ \drawbell{j} \in \drawbellsAt{}(\pose{i}),\ \forall\ s \in [1..\pickupTime{}-1],\\ 
&\forall\ t \in [0..T-1],\ %
\mv{}(\pose{i}, \pose{i}, 0, (t+s)\%T) \geq \pk{}(\pose{i}, \drawbell{j}, t).
\end{align*}
\para{Constraint 4.} Similarly, FLIMM models a $\depositTime{}$ deposit action as a one timestep deposit action followed by $\depositTime{} - 1$ one timestep wait actions. An agent depositing ore into a crusher $\crusher{j} \in \crushersAt{}(\pose{i})$ at pose $\pose{i}$ on timestep $t$ must thus have waited in pose $\pose{i}$ for the preceding $\depositTime{} - 1$ timesteps.
\begin{align*}
&\forall\ \pose{i} \in \allPoses{},\ \forall\ \crusher{j} \in \crushersAt{}(\pose{i}),\ \forall\ s \in [1..\depositTime{}-1],\\ 
&\forall\ t \in [0..T-1],\ %
\mv{}(\pose{i}, \pose{i}, 0, (t-s)\%T) \geq \dpF{}(\pose{i}, \crusher{j}, t).
\end{align*}
Since picks up and depositing ore is usually quick, $\pickupTime{}$ and $\depositTime{}$ is usually 1 (as in our evaluations). When $\pickupTime{} = 1$ and $\depositTime{} = 1$, the range of values $[1..0] = \{\}$ that $s$ takes in Constraints 3 and 4 is empty and the constraint vanishes.

\begin{algorithm}[t]
\small 
\caption{\convFLIMM($\atT{}, \mvT{}, \pkT{}, \dpT{}$)}
\label{alg:ConvFLIMM}
\begin{algorithmic}[1]
\State $n \gets \sum_{\pose{i} \in \allPoses{}} \sum_{\shovelSt{} \in \allShovelSt{}} at(\pose{i}, \shovelSt{}, t)$
\label{ln:count_agents}
\State $(\allAgAt{}, \allAgHas{}) \gets$ two empty $n \times |\ct{}|$ matrices
\State $j \gets 0$
\For{$(\pose{i}, \shovelSt{}) \in \allPoses{} \times \allShovelSt{}$}
\label{ln:iterate_over_pose_shovel_state_pairs}
  \If{$\at{}(\pose{i}, \shovelSt{}, 0) = 1$}
  \label{ln:check_for_agent_in_start_state}
  \State $(\agAt{j0}, \agHas{j0})  \gets (\pose{i}, \shovelSt{})$
  \label{ln:initialize_agent_start_st}
  \State $j \gets j + 1$
  \EndIf
\EndFor 
\For{$(j,t) \in [0..n-1] \times [0..\ct{}-1]$}
  \State $(\agAt{j(t+1)}, \agHas{j(t+1)})  \gets \nextAgState{}(\agAt{jt}, \agHas{jt}, t, \mvT{}, \pkT{}, \dpT{})$
\EndFor 
\State $\agPerm{} \gets$ an empty length $n$ vector
\label{ln:init_agperm}
\For{$j, k \in [0..n-1] \times [0..n-1]$}
  \If{$(\agAt{jT}, \agHas{jT}) = (\agAt{k0}, \agHas{k0})$}
    \State $\agPerm{}(\ag{j}) \gets \ag{k}$
    \label{ln:assign_agperm}
  \EndIf
\EndFor

\State \Return $(\allAgAt{}, \allAgHas{}, \agPerm{})$

\Function{\nextAgState{}}{$\pose{i}, \shovelSt{}, t, \mvT{}, \pkT{}, \dpT{}$}
\For{$\pose{j} \in \adj{}(\pose{i})$}
\label{ln:iterate_through_adj}
  \If{$\mv{}(\pose{i}, \pose{j}, \shovelSt{}, t) = 1$}
  \State \Return $(\pose{j}, \shovelSt{})$
  \label{ln:move_action}
  \EndIf
\EndFor 
\For{$\drawbell{j} \in \drawbellsAt{}(\pose{i})$}
\label{ln:iterate_through_drawbells}
  \If{$\shovelSt{} = 0 \wedge \pk{}(\pose{i}, \drawbell{j}, t) = 1$}
  \State \Return $(\pose{i},1)$
  \label{ln:pickup_action}
  \EndIf
\EndFor 
\For{$\crusher{j} \in \crushersAt{}(\pose{i})$}
\label{ln:iterate_through_crushers}
  \If{$\shovelSt{} = 1 \wedge \dpF{}(\pose{i}, \drawbell{j}, t) = 1$}
  \State \Return $(\pose{i},0)$
  \label{ln:deposit_action}
  \EndIf
\EndFor 
\EndFunction
\end{algorithmic}
\end{algorithm}

\para{Next State Constraints.} These constraints determine the locations of the team of agents on timestep $t + 1 \% \ct{}$ based on the actions that were taken on timestep $t$. 

\para{Constraint 5.} There is an agent at pose $\pose{i}$ with an empty shovel on timestep $t+1 \% \ct{}$ if an agent with an empty shovel moved to $\pose{i}$ or an agent at $\pose{i}$ deposited ore  on timestep $t$. 
\begin{align*}
\forall\ \pose{i} \in & \ \allPoses{}, \forall\ t \in [0..\ct{}-1], \at{}(\pose{i}, 0, t+1 \% \ct{})\\
& = \sum_{\mathclap{\pose{j} \in \adj{}(\pose{i})}} \mv{}(\pose{j}, \pose{i}, 0, t) + \sum_{\mathclap{\crusher{j} \in \crushersAt{}(\pose{i})}} \dpF{}(\pose{i}, \crusher{j}, t).
\end{align*}

\para{Constraint 6.} There is an agent at pose $\pose{i}$ with a full shovel on timestep $t+1 \% \ct{}$ if an agent with a full shovel moved to $\pose{i}$ or an agent at $\pose{i}$ picked up ore on timestep $t$. 
\begin{align*}
\forall\ \pose{i} \in & \ \allPoses{}, \forall\ t \in [0..\ct{}-1],\ \at{}(\pose{i}, 1, t+1\%T)\\ %
& = \sum_{\mathclap{\pose{j} \in \adj{}(\pose{i})}} \mv{}(\pose{j}, \pose{i}, 1, t) + \sum_{\mathclap{\drawbell{j} \in \drawbellsAt(\pose{i})}} \pk{}(\pose{i}, \drawbell{j}, t). 
\end{align*}

\para{Collision Constraints.} These constraints prevent collisions.

\para{Constraint 7.} At most one agent may occupy a pose in any pose conflict set $S \in \poseConflictFamily{}$ on any timestep $t$.
\begin{align*}
\forall\ S \in \poseConflictFamily{}, \forall\ t \in [0..\ct{}-1], \sum_{\mathclap{\pose{i} \in S}} \  \sum_{\mathclap{\shovelSt{} \in \allShovelSt{}}} \at{}(\pose{i}, \shovelSt{}, t) \leq 1.
\end{align*}

\para{Constraint 8.} At most one agent may transition across an edge in any edge conflict set $S \in \edgeConflictFamily{}$ on any timestep $t$.
\begin{align*}
\forall\ S \in \edgeConflictFamily{}, \forall\ t & \in [0..\ct{}-1], \\ &\sum_{(\pose{i}, \pose{j}) \in S} \ \sum_{\mathclap{\shovelSt{} \in \allShovelSt{}}} \mv{}(\pose{i}, \pose{j}, \shovelSt{}, t) \leq 1.
\end{align*}

\para{Draw Ratio Constraint.} This constraint enforces the relative rates at which ore is removed from each drawbell.

\para{Constraint 9.} The ratio between the draw rate from each drawbell $\drawbell{i}$ and the reference drawbell $\drawbell{1}$ must be $\drawRatio{i}$. Recall that $\accessibleFrom{}_i$ be the set of poses from which $\drawbell{i}$ can be accessed.
\begin{align*}
\forall \drawbell{i} \in \allDrawbells{}, \sum_{\mathclap{\pose{j} \in \accessibleFrom{}_i}}\sum_{t=0}^{\ct{}-1} \pk{}(\pose{j}, \drawbell{i}, t) = \drawRatio{i} \sum_{\mathclap{\pose{j} \in \accessibleFrom{}_i}}\sum_{t=0}^{\ct{}-1} \pk{}(\pose{j}, \drawbell{1}, t).
\end{align*}

\para{From Decision Variables to a Plan.} Once FLIMM's ILP is solved, the values assigned to its decision variables encode an optimal solution to the F-BCM.  We translate these decision variable values into a formal plan using the function \convFLIMM{}, shown in Algorithm~\ref{alg:ConvFLIMM}. 

\convFLIMM{} iterates over the variables $\{at(\pose{i}, \shovelSt{}, 0)$ $: \pose{i} \in \allPoses{}, \shovelSt{} \in \shovelSt{}\}$ (Lines~\ref{ln:iterate_over_pose_shovel_state_pairs}-\ref{ln:check_for_agent_in_start_state}). Whenever it encounters a variable with the value 1, it instantiates a new agent $\ag{j}$. Agent $\ag{j}$'s initial pose and shovel state are set to $\agAt{j0} = \pose{i}$ and $\agHas{j0} = \shovelSt{}$ (Line~\ref{ln:initialize_agent_start_st}). Next, \convFLIMM{} determines the action assigned to agent $\ag{j}$ on timestep $t=0$. If:
\begin{itemize}

  \item $\mv{}(\pose{i},\pose{i},\shovelSt{},0)$ is true, agent $\ag{j}$ waits at pose $\pose{i}$. Thus $\agAt{j1}=\pose{i}$ and $\agHas{j1}=\shovelSt{}$. Since each pose has a self-loop, this check occurs in Lines~\ref{ln:iterate_through_adj}-\ref{ln:move_action}.

  \item $\mv(\pose{i},\pose{j},\shovelSt{},0)$ is true for some pose $\pose{j} \in \adj{}(\pose{i})$, agent $\ag{j}$ moves from $\pose{i}$ to $\pose{j}$. Thus $\agAt{j1}=\pose{j}$ and $\agHas{j1}=\shovelSt{}$ (Lines~\ref{ln:iterate_through_adj}-\ref{ln:move_action}).

  \item agent $\ag{j}$'s shovel is empty $\agHas{j0} = 0$ and $\pk{}(\pose{i}, \drawbell{k}, 0) = 1$ for some $\drawbell{k} \in \drawbellsAt{}(\pose{i})$, agent $\ag{j}$ picks up ore from drawbell $\drawbell{k}$. Thus $\agAt{j1} = \pose{i}$ and $\agHas{j1} = 1$ (Lines~\ref{ln:iterate_through_drawbells}-\ref{ln:pickup_action}).

  \item agent $\ag{j}$'s shovel is full $\agHas{j0} = 1$ and $\dpF{}(\pose{i},\crusher{k},0)$ is true for some $\crusher{k} \in \crushersAt{}(\pose{i})$, then $\ag{j}$ deposits ore at crusher $\crusher{k}$. Thus $\agAt{j1}=\pose{i}$ and $\agHas{j1}=0$ (Lines~\ref{ln:iterate_through_crushers}-\ref{ln:deposit_action}).
\end{itemize}

Constraint 5 ensures that exactly one of these conditions holds. ConvFLIMM repeats this procedure to determine the state of agent $\ag{j}$ on timesteps $t=2$ to $T$. After extracting FLIMM's transport plan from its decision variables, ConvFLIMM determines its agent permutation $\agPerm{}$. An agent $\ag{j} \in \allAg{}$ is mapped to the agent $\ag{k}$ if its terminal state $(\agAt{jT}, \agHas{jT})$ matches its initial state $(\agAt{k0}, \agHas{k0})$ (Line~\ref{ln:init_agperm}-\ref{ln:assign_agperm}).

\subsection{The Simple Anytime Mine MAPF (SAMM) solver}

We construct SAMM using our fixed-cycle-length BCM solver, as outlined in Algorithm~\ref{alg:SAMM}. SAMM iteratively runs FLIMM with progressively increasing cycle lengths until the time limit $\timer{}$ is reached, then returns the best plan found. If $\timer{}$ expires while FLIMM is running, it returns $\nullVal{}$.

\begin{algorithm}[t]
\small 
\caption{\SAMM($\OSBCMInstance{}, \timer{}$)}
\label{alg:SAMM}
\begin{algorithmic}[1]
\State $(\bestAllAgAt{}, \bestAllAgHas{}, \bestAgPerm{}, \ct{}) \gets (\nullVal{}, \nullVal{}, \nullVal{},1)$
%
%\State $\ct{} \gets 1$ 
% \Comment{Current number of epochs}
\label{ln:make_num_epochs}

\While{$\timer{}$ has not expired}
\label{ln:checkTimer}
  \State $(\atT{}, \mvT{}, \pkT{}, \dpT{}) \gets \FLIMM{}(\OSBCMInstance{}, \ct{}, \timer{})$
  \label{ln:plan_for_num_epochs}
  \State $(\allAgAt{}, \allAgHas{}, \agPerm{}) \gets \convFLIMM{}(\atT{}, \mvT{}, \pkT{}, \dpT{})$
  \If{$\allAgAt{} = \nullVal{} \vee \textsc{thru}(\allAgAt{}, \allAgHas{}, T) > \textsc{thru}(\bestAllAgAt{}, \bestAllAgHas{}, T)$}
  \label{ln:check_solution_throughput}
    \State $(\bestAllAgAt{}, \bestAllAgHas{}, \bestAgPerm{}) \gets (\allAgAt{}, \allAgHas{}, \agPerm{})$
    \label{ln:update_best_solution}
  \EndIf
  \State $\ct{} \gets \ct{} + 1$
  \label{ln:increment_num_epochs}
\EndWhile
\State \Return $(\bestAllAgAt{}, \bestAllAgHas{}, \bestAgPerm{})$
\label{ln:return_sol}

\Function{\textsc{thru}}{$\allAgAt{}, \allAgHas{}, T$}
  \State \Return $\deposits{}(\allAgAt{}, \allAgHas{})/\ct{}$
\EndFunction
\end{algorithmic}
\end{algorithm}

% The number of variables in FLIMM's ILP scales linearly with the cycle length $\ct{}$. As a result, solving its ILP becomes more computationally demanding as $\ct{}$ increases. Prioritizing smaller values of $\ct{}$ allows SAMM to explore as broad a range of cycle times as possible within the time limit. 

\begin{theorem}
SAMM is eventually optimal.
\end{theorem}

\para{Proof.} Let $(\bestAllAgAt{}, \bestAllAgHas{})$ be the optimal transport plan for the BCM instance $\OSBCMInstance{}$. Let $\bestCt{}$ be the cycle length associated with this plan. Since SAMM enumerates cycle lengths in increasing order, it will eventually call FLIMM on cycle length $\bestCt{}$. FLIMM will return $(\bestAllAgAt{}, \bestAllAgHas{})$ since it is guaranteed to find the optimal plan for any fixed cycle length. $\square$

\smallskip It is usually unnecessary to initialize SAMM with a cycle length of $T=1$ since very short cycle lengths are typically impractical. Our evaluations initialize SAMM with $T=6$. 

\subsection{Accelerating SAMM}

FLIMM constrains each drawbell's draw rate $\drawbell{i}$ to $\drawRatio{i}$ times the reference drawbell's draw rate $\drawbell{1}$. If each ratio $\drawRatio{i}$ is simple (e.g., 3/2), these rates can often be achieved using a plan with a short cycle length. However, complex ratios (e.g., 97/100) may require a long cycle length. FLIMM's worst case runtime grows exponentially with cycle length.

To address this issue, we introduce Flexible Draw Rate FLIMM (FDR-FLIMM), a variant of FLIMM that allows each drawbell's draw rate $\drawbell{i}$ to be higher than the target rate $\drawRatio{i} \drawbell{1}$, relaxing Constraint 9 to read:
\begin{align*}
\forall \drawbell{i} \in \allDrawbells{}, \sum_{\mathclap{\pose{j} \in \accessibleFrom{}_i}}\sum_{t=0}^{\ct{}-1} \pk{}(\pose{j}, \drawbell{i}, t) \geq \drawRatio{i} \sum_{\mathclap{\pose{j} \in \accessibleFrom{}_i}}\sum_{t=0}^{\ct{}-1} \pk{}(\pose{j}, \drawbell{1}, t).
\end{align*}

We enforce the target draw rate for each drawbell at execution time by occasionally skipping pickup actions. Let $\exeDrawRate{}$ be the average rate at which ore has been extracted from drawbell $\drawbell{i}$ from timestep 0 to timestep $t$. On any timestep $t$, if executing a pickup action would cause $\exeDrawRate{}$ to exceed the target rate $\drawRatio{i} \drawbell{1}$, we replace it with $\pickupTime{}$ wait actions.

\begin{theorem}
\label{thm:fdr_flimm_eventuall_optimal}
The throughput of FDR-FLIMM's solution to any fixed cycle length BCM instance $(\OSBCMInstance{}, T)$ is no worse than the throughput of FLIMM's solution.
\end{theorem}

\para{Proof.} FDR-FLIMM is obtained by relaxing one of FLIMM’s constraints. Relaxing a constraint cannot eliminate any feasible solutions. Thus any feasible solution to FLIMM is a feasible solution to FDR-FLIMM. Both FLIMM and FDR-FLIMM maximize the reference draw rate $\drawbell{1}$. It follows that the draw rate $\drawbell{1}$ achieved by FDR-FLIMM is no lower than the draw rate achieved by FLIMM.

The throughput of the solutions produced by both solvers is a multiple of $\drawbell{1}$. FLIMM enforces this relationship through its constraints, while FDR-FLIMM enforces it by selectively skipping pickup actions at execution time. Therefore, the throughput of FDR-FLIMM’s solution is no lower than the throughput of FLIMM’s solution. $\square{}$

Let FDR-SAMM be the variant of SAMM in which FLIMM is replaced by FDR-FLIMM. Theorem~\ref{thm:fdr_flimm_eventuall_optimal} implies that FDR-SAMM is also eventually optimal.

% \begin{theorem}
% The throughput achieved by FDR-FLIMM for any BCM instance $\OSBCMInstance{}$ is at least as high as the throughput achieved by SAMM in the limit of infinite run time.
% \end{theorem}

% \para{Proof.} As runtime tends to infinity, the number of different cycle lengths that both SAMM and FDR-SAMM evaluate tends to infinity. Since FDR-FLIMM weakly dominates FLIMM for any fixed cycle length, it follows that FDR-SAMM achieves throughput at least as high as SAMM.  $\square{}$

% FDR-SAMM is usually more practical than SAMM because it can find a feasible solutions when $\ct{}$ is small. FDR-FLIMM may take longer to solve each fixed cycle length BCM instance than FLIMM, however, because it has to search a larger feasible region. As a result, FDR-SAMM may take longer to arrive at the optimal cycle length $\bestCt{}$.

\begin{algorithm}[t]
\small 
\caption{\SAMMS{}($\OSBCMInstance{} := (\allPoses{}, \allEdges{}, \allDrawbells{}, \allCrushers{}, \drawRatioV{}), \timer{}$)}
\label{alg:SAMMS}
\begin{algorithmic}[1]
\State $\allFamilies{} \gets \partitionDrawbells{}(\allDrawbells{})$
\label{ln:partition_drawbells}
\State $\{\family{1}, \family{2}, \ldots \} \gets \allFamilies{}$
\label{ln:drawbell_families}
\State $\subcycle{1} \gets \genReferenceSubcycle{}(\OSBCMInstance{}, \family{1}, , \timer{})$
\label{ln:gen_reference_subcycle}
\State $
(\allAgAtF{1},\allAgHasF{1},\agPermF{1}) \gets \subcycle{1}$
\State $\agAtSt{} \gets \{\agAt{j0}^{(1)} : j \in [0..n-1]\}$
\label{ln:ref_subcycle_agent_at_st}
\State $\agHasSt{} = \{\agHas{j0}^{(1)} : j \in [0..n-1]\}$
\label{ln:ref_subcycle_agent_at_en}
\For{$\family{i} \in \allFamilies{}\ s.t.\ i \neq 1$}
\label{ln:loop_through_remaining_families}
  \State $\subcycle{i} \gets \genSubcycle{}(\OSBCMInstance{}, \family{i}, \agAtSt{}, \agHasSt{}, \timer{})$
  \label{ln:gen_subcycle}
\EndFor
\State \Return $\composeSubcycles{}(\{\subcycle{i}\}_{\family{i} \in \allFamilies{}})$
\label{ln:compose_subcycles}
\Function{$\genReferenceSubcycle{}$}{$\OSBCMInstance{}, \family{1}, \timer{}$}
\label{ln:gen_reference_subcycle_f}
\State $\reducedBCMInstance{1} \gets \reduceBCM{}(\OSBCMInstance{}, \family{1})$
\label{ln:gen_reduced_BCM_for_family_1}
\State \Return $\SAMM{}(\reducedBCMInstance{1}, \timer{})$
\label{ln:return_reference_subcycle}
\EndFunction

\Function{$\genSubcycle{}$}{$\OSBCMInstance{}, \family{i}, \agAtSt{}, \agHasSt{}, \timer{}$}
\label{ln:gen_subcycle_f}
\State $\reducedBCMInstance{1} \gets \reduceBCM{}(\OSBCMInstance{}, \family{i})$
\label{ln:gen_reduced_BCM_for_family_i}
\State \Return $\FSSAMM{}(\reducedBCMInstance{i}, \agAtSt{}, \agHasSt{}, \timer{})$
\label{ln:return_subcycle_for_family_i}
\EndFunction

\end{algorithmic}
\end{algorithm}

\section{SAMM with Subcycles}

In this section, we introduce SAMMS (SAMM with Subcycles), a scalable solution to the block cave mining problem. We then establish that SAMMS is complete.

The computational complexity of constructing a cyclic plan with an ILP grows exponentially with its cycle length $\ct{}$. Instead of producing a single cyclic plan that services every drawbell, SAMMS constructs a sequence of subcycles $\allSubcycles{} := \{\subcycle{1}, \subcycle{2}, \ldots\}$, each of which services a subset of drawbells. SAMMS constructs a transport plan online by concatenating subcycles from $\allSubcycles{}$. Whenever a subcycle completes, SAMMS selects a new subcycle to execute. SAMMS is shown in Alg. 3. It has four main steps:

\para{Step 1.)} SAMMS partitions the set of drawbells $\allDrawbells{}$ into families using the function $\partitionDrawbells{}$ (Line~\ref{ln:partition_drawbells}). Let the set of drawbell families be $\allFamilies{} := \{\family{1},\family{2},\ldots\}$ (Line~\ref{ln:drawbell_families}). Each drawbell $\drawbell{j}$ belongs to exactly one family.

\para{Step 2.)} Let the reference family $\family{1}$ be the family that contains the reference drawbell $\drawbell{1}$. SAMMS constructs a subcycle $\subcycle{1} := (\allAgAtF{1},\allAgHasF{1},\agPermF{1})$ that services the reference family using the function $\genReferenceSubcycle{}$ (Line~\ref{ln:gen_reference_subcycle}). This subcycle is termed the reference subcycle.

\para{Step 3.)} SAMMS constructs subcycles $\{\subcycle{i}\}_{i \neq 1}$ for the remaining families $\{\family{i}\}_{i\neq 1}$. Let $n$ be the number of agents used by the reference subcycle. Let $(\agAtSt{},\agHasSt{})$ be the state of these agents at $t=0$ (Lines~\ref{ln:ref_subcycle_agent_at_st}-\ref{ln:ref_subcycle_agent_at_en}): 
%
%%\begin{align*}
%&\agAtSt{} = \{\agAt{j0}^{(1)} : j \in [0..n-1]\}\\
%&\agHasSt{} = \{\agHas{j0}^{(1)} : j \in [0..n-1]\}
%\end{align*}
%
\begin{equation*}
\agAtSt{} = \{\agAt{j0}^{(1)} : j \in [0..n-1]\}, \quad \agHasSt{} = \{\agHas{j0}^{(1)} : j \in [0..n-1]\}
\end{equation*}
To ensure that the subcycles that SAMMS constructs can be seamlessly composed into a single global plan, each of these subcycles must: (a) use the same number of agents as the reference subcycle and (b) start and end with their team of agents in the state $(\agAtSt{},\agHasSt{})$. SAMMS constructs these subcycles using the function $\genSubcycle{}$ (Lines~\ref{ln:loop_through_remaining_families}-\ref{ln:gen_subcycle}).

\para{Step 4.)} Finally, SAMMS generates a transport plan online by selecting subcycles to run in real time using the function $\composeSubcycles{}$ (Line~\ref{ln:compose_subcycles}). SAMMS ensures each drawbell's draw rate conforms to the draw ratio $\drawRatio{i}$ by running each subcycle at the appropriate frequency. We now describe the functions associated with each of SAMMS's four steps.

\para{\partitionDrawbells{}.} SAMMS assigns each subcycle $g$ drawbells chosen at random from each drawbell tunnel, where $g$ is a tunable hyperparameter. In block cave mining, overall throughput is typically limited by crusher access. Because drawbells are selected randomly, subcycles typically include drawbells both near and farther away from the main tunnel. This variation in travel distances helps stagger agent arrivals at the crushers, improving their utilization.

\begin{algorithm}[t]
\small
\caption{\reduceBCM{}($\OSBCMInstance{} := (\allPoses{}, \allEdges{}, \allDrawbells{}, \allCrushers{}, \drawRatioV{}), \family{i}$)}
\label{alg:reduce_BCM}
\begin{algorithmic}[1]
\For{$\pose{i} \in \allPoses{}$}
\label{ln:iterate_over_poses}
  \State $\bar{\family{i}}(\pose{i}) \gets \drawbellsAt{}(\pose{i}) \setminus \family{i}$ 
  \label{ln:reduce_drawbells_at}
\EndFor 
\State $\referenceDrawbell{i} \gets \arg \max \{\drawRatio{i} : \drawbell{i} \in \allDrawbells{}$\}.
\label{ln:select_reference_drawbell}
\State $\referenceDrawRatio{i} \gets \max\{\drawRatio{i} : \drawbell{i} \in \allDrawbells{}$\}.
\label{ln:select_reference_draw_ratio}
\State $\reducedDrawRatioV{i} \gets \{\drawRatio{j} / \referenceDrawRatio{i} : \drawbell{j} \in \family{i}\}$
\label{ln:reduce_draw_ratio_v}

\State \Return $(\allPoses{}, \allEdges{}, \family{i}, \allCrushers{}, \reducedDrawRatioV{i})$
\end{algorithmic}
\end{algorithm}

\para{\reduceBCM{}.} Step 2.) and 3.) rely on the function $\reduceBCM{}$, shown in Alg.~\ref{alg:reduce_BCM}. $\reduceBCM{}$ removes every drawbell in an BCM instance $\OSBCMInstance{}$ not in a family $\family{i}$. The resulting reduced BCM instance is denoted $\reducedBCMInstance{i}$.

First, $\reduceBCM{}$ computes the set of drawbells $\bar{\family{i}}(\pose{i})$ accessible from each pose $\pose{j} \in \allPoses{}$ in $\reducedBCMInstance{i}$ by removing every drawbell that is not in $\family{i}$ from the set $\drawbellsAt(\pose{i})$ (Line~\ref{ln:iterate_over_poses}-\ref{ln:reduce_drawbells_at}).

Second, it selects a reference drawbell for $\reducedBCMInstance{i}$. The drawbell in $\family{i}$ with the highest draw rate is chosen as $\family{i}$’s reference drawbell. This drawbell is denoted $\referenceDrawbell{i}$ (Line~\ref{ln:select_reference_drawbell}). Since the reference drawbell $\drawbell{1}$ has the highest draw rate of any drawbell in $\OSBCMInstance{}$, $\drawbell{1}$ is the reference drawbell in $\reducedBCMInstance{1}$.

Third, it re-expresses the draw ratios of the drawbells in $\family{i}$ relative to the new reference drawbell $\referenceDrawbell{i}$. The draw ratio between a drawbell $\drawbell{j} \in \allDrawbells{}$ and the original reference drawbell $\drawbell{1}$ is $\drawRatio{j} := \drawVel{j} / \drawVel{1}$. Thus, the draw ratio between two arbitrary drawbells $\drawbell{j}, \drawbell{k} \in \allDrawbells{}$ is $\drawVel{j} / \drawVel{k} = \drawVel{j} / \drawVel{1} \cdot \drawVel{1} / \drawVel{k} = \drawRatio{j} / \drawRatio{k}$. Let $\referenceDrawRatio{i}$ be the draw ratio between $\family{i}$'s reference drawbell $\referenceDrawbell{i}$ and the original reference drawbell $\drawbell{1}$ in $\OSBCMInstance{}$ (Line~\ref{ln:select_reference_draw_ratio}). The draw ratio between a drawbell $\drawbell{j} \in \family{i}$ and $\referenceDrawbell{i}$ is thus $\drawRatio{j} / \referenceDrawRatio{i}$. The set of rescaled draw ratios $\reducedDrawRatioV{i}$ in $\reducedBCMInstance{i}$ is thus $\reducedDrawRatioV{i} := \{\drawRatio{j} / \referenceDrawRatio{i} : \drawbell{j} \in \family{i}\}$ (Line~\ref{ln:reduce_draw_ratio_v}).

\para{\genReferenceSubcycle{}.} This function is shown in Alg.~\ref{alg:SAMMS} (Lines~\ref{ln:gen_reference_subcycle_f}-\ref{ln:return_reference_subcycle}). Recall that $\family{1}$ is the family containing the original reference drawbell $\drawbell{1}$. $\genReferenceSubcycle{}$ uses $\reduceBCM{}$ to obtain the reduced BCM instance $\reducedBCMInstance{1}$ (Line~\ref{ln:gen_reduced_BCM_for_family_1}). It then uses SAMM to solve $\reducedBCMInstance{1}$ (Line~\ref{ln:return_reference_subcycle}).

\para{\genSubcycle{}.} This function also generates a reduced BCM instance $\OSBCMInstance{i}'$ by applying $\reduceBCM{}$ to $\OSBCMInstance{}$ (Line~\ref{ln:gen_reduced_BCM_for_family_i}). However, it solves $\OSBCMInstance{1}'$ using a modified version of SAMM called Fixed Start SAMM (FS-SAMM) (Line~\ref{ln:return_subcycle_for_family_i}). 

FS-SAMM is constructed by replacing SAMM’s call to FLIMM with a call to a variant of FLIMM, Fixed Start FLIMM (FS-FLIMM). FS-FLIMM considers only solutions to the F-BCM problem that start and end with its agents in the state $(\agAtSt{}, \agHasSt{})$. FS-FLIMM is obtained by augmenting FLIMM with the following constraints.

\para{Constraint 12.} There is an agent in pose $\pose{i} \in \allPoses{}$ with shovel state $\shovelSt{} \in \allShovelSt{}$ on timestep $t=0$ iff there is an agent in that state in $(\agAtSt{}, \agHasSt{})$. 
\begin{align*} 
\forall\ \pose{i} &\in \allPoses{}, \forall\ \shovelSt{} \in \allShovelSt{},\ \at{}(\pose{i}, \shovelSt{}, 0)\\%
&\Leftrightarrow \exists\ j \in [0..n-1]\ s.t.\ (\agAtStF{j}, \agHasStF{j}) = (\pose{i},\shovelSt{}).
\end{align*}

\para{\composeSubcycles{}.} This function constructs a transport plan online by concatenating subcycles from $\allSubcycles{}$. Whenever a subcycle completes, the function selects the next subcycle to execute. Because all subcycles share the same start and end state, they can be seamlessly composed in any order.

\composeSubcycles{} ensures that the draw rates of all drawbells conform to the draw ratios in $\drawRatioV{}$ by adjusting how frequently each subcycle is executed. Let $\frequency{i}$ be subcycle $\subcycle{i}$'s execution frequency. Let $\pickup(\referenceDrawbell{i})$ be the amount of ore that each execution of $\subcycle{i}$ removes from its reference drawbell $\referenceDrawbell{i}$. The draw rate from $\referenceDrawbell{i}$ is thus $\frequency{i} \cdot \pickup(\referenceDrawbell{i})$.

The required draw ratio between $\referenceDrawbell{i}$ and subcycle $\subcycle{1}$'s reference drawbell $\referenceDrawbell{1} = \drawbell{1}$ is $\referenceDrawRatio{i}$. To produce this draw ratio, \composeSubcycles{} must ensure that:
\begin{equation*} 
\frequency{i} \cdot \pickup{}(\referenceDrawbell{i}) = \frequency{1} \cdot \pickup{}(\referenceDrawbell{1}) \cdot \referenceDrawRatio{i} 
\end{equation*}
Thus, $\subcycle{i}$ must be executed $\executionRatio{i}$ times for every execution of $\subcycle{1}$ where:
\begin{equation*}
\executionRatio{i} := \frac{\frequency{i}}{\frequency{1}} \;=\; \frac{\pickup(\referenceDrawbell{1}) \cdot \referenceDrawRatio{i}}{\pickup(\referenceDrawbell{i})}
\end{equation*}
We term $\executionRatio{i}$ subcycle $\subcycle{i}$'s execution ratio. Let the set of all execution ratios be $\allExecutionRatios{} := \{\executionRatio{i} : \subcycle{i} \in \allSubcycles{}\}$.

In practice, achieving the exact execution ratio $\executionRatio{i}$ that  each subcycle $\{\subcycle{i}\}_{i \neq 1}$ requires is difficult when these ratios are not simple. Consequently, $\composeSubcycles{}$ is designed to achieve asymptotic feasibility. As the plan length increases, $\composeSubcycles{}$'s draw ratio constraint violations converge to zero at the rate $O(1/n)$, where $n$ is the number of subcycles that have been executed. $\composeSubcycles{}$ is shown in Alg.~\ref{alg:compose_subcycles}.

$\composeSubcycles{}$ continously loops over the subcycles in $\allSubcycles{}$. When it reaches each subcycle $\subcycle{i} \in \allSubcycles{}$, it runs $\subcycle{i}$ unless doing so would cause $\subcycle{i}$'s realized execution ratio to exceed its target execution ratio (Lines~\ref{ln:reach_subcycle}-\ref{ln:measure_frequency}).

\subsection{Analysis}

Let $L$ be the number of times that \composeSubcycles{}'s inner loop (Lines~\ref{ln:reach_subcycle}-\ref{ln:measure_frequency}) has been run.

\begin{theorem} 
SAMMS can generate an asymptotically feasible solution for any BCM instance $\OSBCMInstance{}$. After \composeSubcycles{} loops $L$ times, each drawbell's draw rate  within $O(1/L)$ of the target draw rate.
\end{theorem}

\para{Proof.} The realized draw ratio between a drawbell $\drawbell{j} \in \family{i}$ and the reference drawbell $\drawbell{1}$ is directly proportional to the realized execution ratio between $\family{i}$ and $\family{1}$. Thus, if the realized execution ratio of $\family{i}$ converges to $\executionRatio{i}$ at rate $O(1/L)$, the realized draw ratio of drawbell $\drawbell{j}$ converges to $\drawRatio{j}$ at the same asymptotic rate. After $L$ loops, each subcycle $\subcycle{k} \in \allSubcycles{}$ has been executed exactly $\lfloor \executionRatio{k} \cdot L \rfloor$ times. Since $\executionRatio{1} = 1$, the realized execution ratio of $\subcycle{i}$ is $\lfloor \executionRatio{i} \cdot L \rfloor / L$. The error $\executionRatio{i} - \lfloor \executionRatio{i} \cdot L \rfloor / L$ converges to $0$ at rate $O(1/L)$. $\square$

\begin{algorithm}[t]
\small 
\caption{\composeSubcycles{}($\allSubcycles{}, \allExecutionRatios{}$)}
\label{alg:compose_subcycles}
\begin{algorithmic}[1]
\For{$\subcycle{i} \in \allSubcycles{}$}
  \State $\frequency{i} \gets 0$
\EndFor
\While{$true$}
\For{$\subcycle{i} \in \allSubcycles{}$}
\label{ln:reach_subcycle}
  \If{$(\frequency{i} + 1) / \frequency{1} \leq \executionRatio{i}$}
    \State $\run{}(\subcycle{i})$
    \State $\frequency{i} \gets \frequency{i} + 1$
    \label{ln:measure_frequency}
  \EndIf
\EndFor
\EndWhile
\end{algorithmic}
\end{algorithm}

\begin{table*}[t]
  \centering
  \label{tab:samm_samms}
  \caption{SAMM vs.\ SAMMS throughput and runtime across scenarios.}
  \vspace{-3mm}
  \begin{tabular}{r r r r r r r}
    \toprule
    \textbf{Scenario} & \textbf{Drawbells} & \textbf{Crushers} & \textbf{SAMM Throughput }& \textbf{SAMM Runtime (s)} & \textbf{SAMMS Throughput} & \textbf{SAMMS Runtime (s)} \\
    \midrule
    1 & 8  & 4  & 0.89 & 11.0  & 0.56 & 0.96\\
    2 & 14 & 7  & 1.56 & 49.1  & 0.92 & 1.69  \\
    3 & 16 & 8  & 1.78 & 57.8  & 1.10 & 2.92  \\
    4 & 18 & 4  & N/A & 60 & 0.40  & 11.28\\
    5 & 24 & 4  & N/A  & 60 & 0.34  & 55.8 \\
    6 & 36 & 6  & N/A  & 60 & 0.59  & 13.9 \\
    7 & 48 & 8  & N/A  & 60 & 0.73  & 34.0 \\
    \bottomrule
  \end{tabular}
    \vspace{-6mm}
\end{table*}

\section{Experimental Results}
We evaluate SAMM and SAMMS from two points of view: (a)~the computational time, and (b)~solution quality, expressed as throughput, i.e. the total amount of ore deposited.

\para{Setup.} SAMM and SAMMS are evaluated on five block cave mine scenarios. These scenarios were chosen to have a good mix of drawbells and crushers, maximizing the potential effects of optimization. SAMM and SAMMS were given a time limit of 60 seconds in each experiment. This time limit is realistic since embeddings are computed offline. Both SAMM and SAMMs were set to terminate after two runs of FLIMM without any improvement. Pickup and deposit times were both set to one timestep. The scenarios~\cite{github_scenarios} are produced by a benchmark generator whose parameters are calibrated from layout statistics reported in~\cite{Gomez2020LayoutsBlockCaving}. We implement\cite{leet2026sammcode} Alg. \ref{alg:SAMM} and \ref{alg:SAMMS} in Python 3.11~\cite{python_3_11}. We represent the layout graph with the NetworkX~\cite{networkx} library, and solve FLIMM as an ILP using Gurobi~\cite{gurobi}. Our implementation and the code to run our evaluations is publicly available at~\cite{leet2026sammcode}. Each evaluation was performed on a Intel Core Ultra 9 275HX $\times$ 24 CPU,  a NVIDIA GeForce RTX 5080 Laptop GPU and 32 GiB of RAM running Ubuntu 24.04.3 LTS.

\subsection{Results.} 
% Requires \usepackage{booktabs}

The goal of our evaluation is to determine whether SAMM and SAMMS can scale to scenarios representative of industrial use cases. We ran SAMM and SAMMS 5 times on 7 scenarios of increasing complexity. Table~\ref{tab:samm_samms} shows each solvers: (i) mean throughput, measured across the runs where it terminated successfully, and (ii) mean runtime, where a failed run had a mean runtime of 60sec. SAMMS solved 5/5 runs on each scenario except for scenarios 5 and 7, where it solved 4/5 runs. It consistently achieved runtimes an order of magnitude faster than SAMM by decomposing the BCM problem into smaller, independent subproblems. SAMM, however, was not able to scale to the two larger scenarios. Notably, throughput is always relatively low, even in scenarios with a large number of crushers. In the tight mine geometries used in our evaluations, maneuvering agents into and out of a crusher is time-consuming. Consequently, whenever an agent deposits ore into a crusher, the crusher is effectively occupied or an extended period of time, even though a deposit action only takes a single timestep.

\section{Conclusions}
In this paper, we formalize the \emph{Block Cave Mining (BCM)} problem, which focuses on coordinating autonomous vehicles in underground mines. In order to solve the problem, we propose SAMM, an eventually optimal anytime solver, which jointly solves task assignment, scheduling, and path planning via MILP. To address scalability limitations inherent in MILP approaches, we also present SAMMS, which decomposes the planning into shorter subcycles, trading optimality for computational efficiency. Experimental results on real-world mine scenarios demonstrate that SAMMS can be applied to industrial-scale BCM instances.

\bibliographystyle{IEEEtran}
\bibliography{IEEEabrv,references}

\end{document}